\def\year{2018}\relax
\documentclass[letterpaper]{article} 
\usepackage{aaai18}  
\usepackage{times}  
\usepackage{url}  
\usepackage{graphicx}  
\usepackage{booktabs}
\usepackage{multirow}
\usepackage[hyperindex,breaklinks]{hyperref}
\begin{document}
%
\title{Efficient Large-Scale Multi-Modal Classification}
\author{Douwe Kiela, Edouard Grave, Armand Joulin and Tomas Mikolov\\
Facebook AI Research\\
\{dkiela,egrave,ajoulin,tmikolov\}@fb.com
}
\maketitle
\begin{abstract}
While the incipient internet was largely text-based, the modern digital world is becoming increasingly multi-modal. Here, we examine multi-modal classification where one modality is discrete, e.g. text, and the other is continuous, e.g. visual representations transferred from a convolutional neural network. In particular, we focus on scenarios where we have to be able to classify large quantities of data quickly. We investigate various methods for performing multi-modal fusion and analyze their trade-offs in terms of classification accuracy and computational efficiency. Our findings indicate that the inclusion of continuous information improves performance over text-only on a range of multi-modal classification tasks, even with simple fusion methods. In addition, we experiment with discretizing the continuous features in order to speed up and simplify the fusion process even further. Our results show that fusion with discretized features outperforms text-only classification, at a fraction of the computational cost of full multi-modal fusion, with the additional benefit of improved interpretability.
\end{abstract}

\noindent Text classification is one of the core problems in machine learning and natural language processing \cite{Borko:1963acm,Sebastiani:2002acm}. It plays a crucial role in important tasks ranging from document retrieval and categorization to sentiment and topic classification \cite{Deerwester:1990jasis,Joachims:1998ecml,Pang:2008ftir}. However, while the incipient Web was largely text-based, the recent decade has seen a surge in multi-modal content: billions of images and videos are posted and shared online every single day. That is, text is either replaced as the dominant modality, as is the case with Instagram posts or YouTube videos, or it is augmented with non-textual content, as with most of today's web pages. This makes multi-modal classification an important problem.

Here, we examine the task of multi-modal classification using neural networks. We are primarily interested in two questions: what is the best way to combine (i.e., fuse) data from different modalities, and how can we do so in the most efficient manner? We examine various efficient multi-modal fusion methods and investigate ways to speed up the fusion process. In particular, we explore discretizing the continuous features, which leads to much faster training and requires less storage, yet is still able to benefit from the inclusion of multi-modal information. To the best of our knowledge, this work constitutes the first attempt to examine the accuracy/speed trade-off in multi-modal classification; and the first to directly show the value of discretized features in this particular task.

If current trends continue, the Web will become increasingly multi-modal, making the question of multi-modal classification ever more pertinent. At the same time, as the Web keeps growing, we have to be able to efficiently handle ever larger quantities of data, making it important to focus on machine learning methods that can be applied to large-scale scenarios. This work aims to examine these two questions together.

Our contributions are as follows. First, we compare various multi-modal fusion methods, examine their trade-offs, and show that simpler models are often desirable. Second, we experiment with discretizing continuous features in order to speed up and simplify the fusion process even further. Third, we examine learned representations for discretized features and show that they yield interpretability as a beneficial side effect. The work reported here constitutes a solid and scalable baseline for other approaches to follow; our investigation of discretized features shows how multi-modal classification does not necessarily imply a large performance penalty and is feasible in large-scale scenarios.


\section{Related work}

\paragraph{Text classification.}

Neural network-based methods have become increasingly popular for text classification \cite{Socher:2011emnlp,Wang:2012acl}. Recent work has used neural networks for text classification either at a sentence \cite{Kim:2014iclr,Hill:2016iclr} or full document \cite{Le:2014icml,Baker:2016coling,Joulin:2016arxiv} level. Many core NLP tasks are essentially text classification, from tweets \cite{Sriram:2010sigir} to reviews to spam. Even though there has been extensive work on feature engineering for text classification \cite{Chen:2009esa}, modern approaches often make use of word embeddings \cite{Mikolov:2013nips} or sentence representations \cite{Kiros:2015nips} learned from a large corpus in an unsupervised fashion.

\paragraph{Fusion strategies.}

Multi-modal fusion, or the integration of input from various modalities, is an important topic in the field of multimedia analysis \cite{Wu:2004acm,Atrey:2010ms}. The question of fusion has been explored in a variety of tasks, from audio-visual speech recognition \cite{Potamianos:2003ieee} to multi-sensor management \cite{Zhao:2003csur} and face recognition \cite{Xiong:2002if}. Much of this research has focused on the combination of two or more continuous modalities. Here, we are specifically interested in the fusion of discrete textual input with another, continuous, modality.

\paragraph{Multi-modal NLP.}

The usage of non-textual information in natural language processing \cite{Mooney:2008aaai} has become increasingly popular. On the one hand, there has been a lot of interest in cross-modal applications, such as image annotation \cite{Weston:2011ijcai}, image captioning \cite{Bernardi:2016jair}, mapping images to text or vice versa \cite{Frome:2013nips,Socher:2013nips,Lazaridou:2014acl} and visual question answering \cite{Antol:2015iccv,Fukui:2016arxiv}. On the other hand, multi-modal fusion has been extensively explored in the context of grounded representation learning for lexical semantics \cite{Bruni:2014jair,Kiela:2014emnlp,Lazaridou:2015iclr}. While much of this work has focused on vision \cite{Baroni:2016llc}, other perceptual modalities modalities \cite{Lopopolo:2015iwcs,Kiela:2015emnlp,Kiela:2015acl} have also been explored, as well as robotics \cite{Mei:2016aaai}, videos \cite{Regneri:2013tacl} and games \cite{Branavan:2012jair,Narasimhan:2015emnlp}. This work is similar in spirit to \cite{Bruni:2014jair}, in that we explore fusion techniques. However, similarly to \cite{Lazaridou:2015iclr}, we learn how to integrate the multi-modal inputs, and use transferred representations as in \cite{Kiela:2014emnlp}.

\paragraph{Multi-modal deep learning.}

Our work relates to previous work on integrating information from multiple modalities in neural networks \cite{Ngiam:2011icml,Srivastava:2012nips,Kiros:2014icml}. Here, we enhance a  well-known neural network architecture for efficient text classification \cite{Joulin:2016arxiv} with the ability to include continuous information, and explore methods for combining multi-modal features. The works of \cite{Arevalo:2017iclr} and \cite{Fukui:2016arxiv}, explore complex gating mechanisms and compact bilinear pooling as multi-modal fusion methods. In order to obtain visual representations, we transfer continuous features from neural networks trained on other tasks (in this case ImageNet), as has been shown to work well for a wide variety of tasks \cite{Oquab:2014cvpr,Razavian:2014cvpr}.

\begin{table*}[ht]
	\centering
	\begin{tabular}{l|rr|rr|rr}
		\toprule
		Dataset & \#Train & \#Words & \#Valid & \#Words & \#Test & \#Words\\\midrule
		Food101 & 58,131 & 98,365,392 & 6,452 & 10,893,597 & 21,519 & 36,955,182\\
		MM-IMDB & 15,552 & 2,564,734 & 2608 & 425,863 & 7799 & 1,266,681\\
		FlickrTag & 70,243,104 & 1,134,118,808 & 656,687 & 10,100,945 & 621,444 & 9,913,566\\
		FlickrTag-1 & 7,166,110 & 92,651,036 & 48,048 & 682,663 & 48,471 & 672,900\\
		\bottomrule
	\end{tabular}
	\caption{\label{table:evaluation-numbers}Evaluation datasets with their quantitative properties.}
\end{table*}

\section{Evaluation}
\label{sec:evaluations}

Surprisingly, there are not many large-scale multi-modal classification datasets available. We evaluate on three datasets that are large enough to examine accuracy/speed trade-offs in a meaningful way. Two of our datasets (Food101 and MM-IMDB) are medium-sized; while the third dataset (FlickrTag) is very large by today's standards. The quantitative properties of the respective datasets are shown in Table \ref{table:evaluation-numbers} and they are described in more detail in what follows.

\subsection{Food101}

The UPMC Food101 dataset \cite{Wang:2015icme} contains web pages with textual recipe descriptions for 101 food labels automatically retrieved online. Each page was matched with a single image, where the images were obtained by querying Google Image Search for the given category. Examples of food labels are \emph{Filet Mignon}, \emph{Pad Thai}, \emph{Breakfast Burrito} and \emph{Spaghetti Bolognese}. The web pages were processed with html2text\footnote{https://pypi.python.org/pypi/html2text} to obtain the raw text.

\subsection{MM-IMDB}

The recently introduced MM-IMDB dataset \cite{Arevalo:2017iclr} contains movie plot outlines and movie posters. The objective is to classify the movie by genre. This is a multi-label prediction problem, i.e., one movie can have multiple genres. The dataset was specifically introduced to address the lack of multi-modal classification datasets.

\subsection{FlickrTag and FlickrTag-1}

We use the FlickrTag dataset based on the massive YFCC100M Flickr dataset of \cite{Thomee:2016acm} that was used in \cite{Joulin:2016arxiv}. The dataset consists of Flickr photographs together (in most, but not all cases) with short user-provided captions. The objective is to predict the user-provided tags that belong to the photograph. This is a very large-scale dataset, so we perform the multi-modal fusion operator and speed-versus-accuracy studies on a subset (specifically, the first shard, which corresponds to one-tenth of the full dataset) for those studies, which we denote FlickrTag-1. We show that the inclusion of discretized features yields classification accuracy improvements with respect to text on the whole dataset.

\section{Approach}

As a starting point, we take the highly efficient text classification approach of FastText \cite{Joulin:2016arxiv}. To ensure a fair comparison, we enhance that model with the capability to handle continuous or discretized features. Specifically, we use 2048-dimensional continuous features that were obtained by transferring the pre-softmax layer of a 152-layer ResNet \cite{He:2016cvpr} trained on the ImageNet classification task. In the case of the large-scale FlickrTag datasets, we use ResNet-34 features (of 512 dimensions). It has been shown that convolutional network features can be transferred successfully to a variety of tasks \cite{Razavian:2014cvpr} and we take the same approach here. We explore a variety of models and experiment with discretization.

The scenario of multi-modal classification certainly admits, or even invites, highly sophisticated models. In our case, however, we also have to take into account efficiency, so we want to focus on models that are simple and efficient enough to handle large-scale datasets, while obtaining improved performance over our baselines. We experiment with a comprehensive set of models, listed below in increasing order of complexity.

In all cases, given a set of N documents, the objective is to minimize the negative log likelihood over the classes:

\begin{equation}
	- \frac{1}{N} \sum^N_{n=1} \log (softmax(o(x_n), y_n)),
\end{equation}

\noindent where $o$ is the network's output, $x_n$ is the multi-modal input and $y_n$ is the label.


\subsection{Baselines}

\paragraph{Text} The first baseline consists of FastText \cite{Joulin:2016arxiv}, a library for highly efficient word representation learning and sentence classification. FastText is trained asynchronously on multiple CPUs using stochastic gradient descent and a learning rate that linearly decays with the amount of words. It yields competitive performance with more sophisticated text classification approaches, while being much more efficient. That is, we ignore the visual signal altogether and only use textual information, i.e.,

\[o(x_n) = W U x^t_n,\]

\noindent where $W$ and $U$ are weight matrices and $x^t_n$ is the normalized bag of textual features representation.

\paragraph{Continuous}

The second baseline consists of training a classifier only on top of the transferred ResNet features \cite{He:2016cvpr}. That is, we ignore the textual information and only use the visual input, i.e.,

\[ o(x_n) = W V x^v_n,\]

\noindent where $W$ and $V$ are weight matrices and $x^v_n$ consists of the ResNet features, normalized to unit length.

\subsection{Continuous Multi-Modal Models}

\paragraph{Additive} We combine the information from both modalities using component-wise addition, i.e.,

\[o(x_n) = W (U x^t_n + V x^v_n).\]


\paragraph{Max-pooling} We combine the information from both modalities using the component-wise maximum, i.e.,

\[o(x_n) = W \max(U x^t_n, V x^v_n).\]

\paragraph{Gated} We allow one modality to ``gate'' or ``attend'' over the other modality, via a sigmoid non-linearity, i.e., 

\[o(x_n) = W (\sigma(U x^t_n) * V x^v_n),\]

or alternatively,

\[o(x_n) = W (U x^t_n * \sigma(V x^v_n)).\]

\noindent One can think of this approach as performing attention from one modality over the other. It is a conceptually similar simplification of multi-modal gated units, introduced in \cite{Arevalo:2017iclr}. The modality to be gated is a hyperparameter (see below).

\paragraph{Bilinear} Finally, to fully capture any associations between the two different modalities, we examine a bilinear model, i.e.,

\[o(x_n) = W (U x^t_n \otimes V x^v_n).\]

\noindent This approach can be thought of as a simpler version of the more complex multi-modal bilinear pooling introduced by \cite{Fukui:2016arxiv}. We also experiment with a method where we introduce a gating non-linearity into the bilinear model, which we call \textbf{Bilinear-Gated}.

\begin{table*}[t]
	\centering
	\begin{tabular}{lllll}
		\toprule
		& \textbf{Model} & \textbf{Food101} & \textbf{MM-IMDB} & \textbf{FlickrTag-1}\\
		\midrule
		\multirow{3}{*}{Previous work} & \cite{Wang:2015icme} & 85.1 & --- & ---\\
		& \cite{Arevalo:2017iclr}-GMU & --- & \textbf{63.0} & ---\\
		& \cite{Arevalo:2017iclr}-AVG & --- & 61.5 & ---\\
		\midrule
		\multirow{2}{*}{Baselines} & FastText & 88.0 $\pm$ 0.1 & 58.8 $\pm$ 0.1 & 23.0 $\pm$ 0.0\\
		& Continuous & 56.7 $\pm$ 0.2 & 49.3 $\pm$ 0.0 & 12.4 $\pm$ 0.0\\
		\midrule
		\multirow{5}{*}{Continuous} & Additive & 90.4 $\pm$ 0.1 & 61.0 $\pm$ 0.0 & 26.8 $\pm$ 0.0\\
		& Max-pooling & 90.5 $\pm$ 0.1 & 62.2 $\pm$ 0.1 & 26.9 $\pm$ 0.0\\
		& Gated & 90.1 $\pm$ 0.1 & 61.8 $\pm$ 0.1 & 27.7 $\pm$ 0.0\\
		& Bilinear & 88.1 $\pm$ 0.3 & 61.5 $\pm$ 0.1 & 27.8 $\pm$ 0.0\\
		& Bilinear-gated & \textbf{90.8 $\pm$ 0.1} & \textbf{62.3 $\pm$ 0.2} & \textbf{28.6 $\pm$ 0.0}\\\midrule
		\multirow{2}{*}{Discretized} & PQ (n=4, k=256) & 89.5 $\pm$ 0.1  &  60.5 $\pm$ 0.1  &  25.6 $\pm$ 0.1\\
		& RSPQ (n=4, k=256, r=4) & 89.8 $\pm$ 0.0  &  60.7 $\pm$ 0.1  &  26.1 $\pm$ 0.1\\
		\bottomrule
	\end{tabular}
	\caption{\label{table:results}Accuracy (averaged over 5 runs) of continuous and discretized multi-modal models, compared to baselines.}
\end{table*}

\subsection{Discretized Multi-Modal Models}

A downside of continuous models is that they require an expensive matrix-vector multiplication $Vx^v_i$ and storing large matrices of floating point numbers requires a lot of space. While the ResNet features used in these experiments consist of a relatively small number of components, these can easily run into the tens of thousands: consider e.g. combinations of SIFT and Fisher vectors used in state-of-the-art computer vision applications \cite{Perronnin:2015cvpr}. Hence, we experiment with discretizing the continuous features, where we convert the continuous features to a discrete sequence of tokens, which can be treated as if they are special tokens, which we normalize separately, and used in the standard FastText setup. This is a simple, computationally less intensive solution. Discretized features also obviously require less storage.

In particular, we investigate product quantization (PQ) \cite{Jegou:2011pami}, where we divide the continuous vector into subvectors of equal size, and then perform k-means clustering on each of the subvectors. For each image, we subsequently determine the closest centroid for each of its subwords, which is combined with the subvector index in order to obtain a discretized vector. For example, a $100$-dim continuous vector $x^v_i$ may be divided into ten $10$-dimensional subvectors $s_i$. Let $N(s_i)$ denote the index of the nearest centroid for $s_i$. The discretized representation of $v$ is then given as $\langle (1, N(s_1)), (2, N(s_2)), .., (10, N(s_{10}))\rangle$. We include these tokens in the text and treat them as if they were special tokens, in the standard fastText model, i.e.,

\[o(x_n) = W (U x^t_n + \alpha U x^d_n).\]

\noindent where $x^d_n$ are the discretized features and $\alpha$ is a reweighting hyperparameter. We normalize $x^t_n$ and $x^d_n$ independently. As we can see, the discretized models are closely related to the additive model, except that they use the same weight matrix $U$ with the discretized features used as ``words'' in the text.

While PQ is great for compressing information into a discretized sequence, it does impose hard boundaries on subvectors, which means that overlapping semantic content that is shared between subvectors may be lost. Hence, we introduce a novel quantization method, called random sample product quantization (RSPQ), in order to maintain (at least some) overlapping semantic information. In RSPQ, the process is the same as in PQ, except we perform PQ over $r$ repetitions of random permutations of $x^v_i$. In both cases, we treat the discretized features as if they are reweighted special tokens included in the textual data and run standard fastText.

\subsection{Model complexity}

There are various trade-offs at stake between these models. The additive, max-pooling and gated models are simplest and result in a hidden layer of the same size as with the normal FastText. The computational complexity of the linear classifier is thus $O(H K)$, where $K$ is the number of classes and $H$ is the size of $Ux^t$ and $Vx^v$. The max-pooling and gating models are slightly more complicated than the additive one, requiring an extra operation. For the bilinear model the complexity amounts to $O(H^2 K)$. Thus, the bilinear model is by far the most expensive to compute. The additive model has the benefit that it does not strictly require a continuous input at all times.

\subsection{Hyperparameters and training}

In all experiments, the model is tuned on the validation set. We tried the following hyperparameters: a learning rate in~$\{ 0.1, 0.25, 0.5, 1.0, 2.0 \}$, a number of epochs in~$\{5, 10, 20\}$, a reweighting parameter in~$\{0.01, 0.02, 0.05, 0.1, 0.2, 0.5\}$ and an embedding dimensionality of either $20$ or $100$. These hyperparameters were sweeped using grid search and we used a softmax loss. For other hyperparameters, such as the number of threads in the parallel optimization and the minimum word count, we fixed their values to standard values in FastText (4 threads, minimum count of 1, respectively), since we found that these did not impact classification accuracy. In the case of the gated and bilinear-gated models, the modality used to serve as a gate over the other modality is treated as a hyperparameter as well.

\section{Results}

The results of the comparison may be found in Table \ref{table:results}. We compare the continuous and discretized multi-modal models against the text-only FastText model and to the continuous features-only model. We also include results from \cite{Wang:2015icme} on Food-101, where they used TF-IDF features for text and a deep convolutional neural network features for images, as well as results from \cite{Arevalo:2017iclr} for Gated Multimodal Units (GMU) and their AVG\_Probs model. GMUs are a substantially more complicated model architecture than any of our relatively simple fusion methods, so this study is a good test of their capability. We note that in the case of Food101, our methods work considerably better than previously reported results. For MM-IMDB, the continuous multi-modal models perform very close to the GMU model and outperform the AVG\_Probs method, while being simpler and computationally more efficient.

We observe that multi-modal models always outperform standard FastText and the continuous-only approach, disregarding the particular type of fusion. This shows that the inclusion of multi-modal information (at least in these types of classification tasks) always helps and that making use of multi-modal information, where available, will lead to increased performance. FastText outperforms the continuous-only method on all datasets, which indicates that text plays a big role in these tasks, and that it is relatively more important than the visual information.

If we examine the continuous multi-modal models, we see that the bilinear-gated model is the clear winner: it outperforms all other methods on all three tasks. It is however also the most complicated model, and as a result is less efficient. We found that placing the gating non-linearity on the text led to the best performance on Food101 and MM-IMDB, while placing it on the visual modality led to the best performance on FlickrTag. It is interesting to observe that the more complicated gated model, as well as the non-gated bilinear model, do not necessarily outperform the simpler additive and max-pooling models. In fact, the performance of these much simpler models is not too far removed from the best scores. The take-away message appears to be: if you care more about accuracy, use the bilinear method with gating; if you care more about speed, use a simple model like the additive or max-pooling one, which have the additional potential benefit that they do not necessarily require the presence of continuous information if none is available.

\begin{table}[t]
	\centering
	\begin{tabular}{lr}
		\toprule
		\textbf{Model} & \textbf{Train time (FlickrTag-1)}\\
		\midrule
		FastText    &  0h01m\\\midrule
		Additive      &  0h39m\\
		Max-pooling & 0h39m\\
		Gated     &  0h40m\\
		Bilinear &  1h04m\\
		Bilinear-Gated &  1h06m\\\midrule
		PQ    &  0h01m\\
		RSPQ    &  0h02m\\
		\bottomrule
	\end{tabular}
	\caption{\label{table:results-train-time}Training time on FlickrTag-1}
\end{table}

\begin{table}[t]
	\centering
	\begin{tabular}{lr}
		\toprule
		\textbf{Model} & \textbf{FlickrTag}\\
		\midrule
		FastText & 36.7\\\midrule
		PQ (n=4, k=256) & 38.9\\
		RSPQ (n=4, k=256, r=4) & 39.4\\
		\bottomrule
	\end{tabular}
	\caption{\label{table:results-flickrtag}Performance on full FlickrTag.}
\end{table}

\subsection{Speed}

We can draw inspiration from the fact that the additive model performed reasonably well: if speed is essential---if necessary at the expense of some accuracy---the discretized models are an obvious choice to further simplify and speed up the model. Even though they outperform standard FastText by a large margin, as shown in Table \ref{table:results}, they only come with a minor performance impact. Table \ref{table:results-train-time} shows the training times for the various models on the FlickrTag-1 dataset: while the bilinear models take around one hour to train (recall that this constitutes only the first shard of the full dataset); the discretized methods, similarly to FastText, only take around one minute. If we scale up to the full FlickrTag dataset, Table \ref{table:results-flickrtag} shows that the discretized models substantially outperform standard FastText. An increase of 2.7\% in accuracy, as seen from FastText to RSPQ, represents having an additional $16778$ test set documents correctly classified using that model, which is a non-negligible amount.

\begin{table}[t]
	\centering
	\begin{tabular}{lll}
		\toprule
		\textbf{$\_\_q\_\_0\_253$} & \textbf{$\_\_q\_\_0\_13$} & \textbf{$\_\_q\_\_1\_253$}\\
		\midrule
donuts 0.987 & cr\`eme 0.933 & oishii 0.905\\
doughnuts 0.981 & ramekins 0.928 & shoga 0.885\\
donut 0.980 & brulee 0.925 & tenkasu 0.884\\
doughnut 0.979 & br\^ul\'ee 0.916 & octopus 0.883\\
donuts? 0.939 & custards 0.916 & aonori 0.881\\
		\bottomrule
	\end{tabular}
	\caption{\label{table:results-interpretability}Examples of nearest neighbors for quantized features in Food101.}
\end{table}

\subsection{Interpretability}

An interesting side effect of the discretized multi-modal methods is that they allow us to examine the nearest neighbors of the quantized features: if a particular feature corresponds to something that looks like a donut, for example, then its embedding should be close to words related to \emph{donut}. Indeed, as Table \ref{table:results-interpretability} shows, we can find clearly identifiable clusters, e.g. for donuts, cr\`eme br\^ul\'ee and certain types of Japanese food. Interpretability is an important but often overlooked aspect of classification models: we show that a simple and efficient method, that outperforms text-only methods by a large margin, yields the additional benefit that it allows for the interpretation of the visual features that a classifier picks up on---something that is difficult to achieve with standard convolutional features.

\section{Conclusion \& Outlook}

The internet is becoming increasingly multi-modal, which makes the task of multi-modal classification ever more pertinent. In order to be able to handle large quantities of data, we need efficient models for large-scale multi-modal classification. In this work, we examined these two questions together. First, we compared various multi-modal fusion methods and found a bilinear-gated model to achieve the highest accuracy, while the simpler additive and max-pooling models yielded reasonably high accuracy at higher speed. Second, we showed that the model can be speeded up even further by introducing discretized multi-modal features. Lastly, we showed that this method yields the additional benefit of interpretability, where we can examine what the multi-modal model picks up on when making its classification decision. We hope that this work can serve as a useful baseline for further work in multi-modal classification.

\section{Acknowledgments}

We thank the reviewers for their comments and our colleagues at FAIR for their feedback and support.

\bibliography{references}

\begin{thebibliography}{}

\bibitem[\protect\citeauthoryear{Antol \bgroup et al\mbox.\egroup
  }{2015}]{Antol:2015iccv}
Antol, S.; Agrawal, A.; Lu, J.; Mitchell, M.; Batra, D.; Lawrence~Zitnick, C.;
  and Parikh, D.
\newblock 2015.
\newblock {VQA}: Visual question answering.
\newblock In {\em Proceedings of the IEEE International Conference on Computer
  Vision},  2425--2433.

\bibitem[\protect\citeauthoryear{Arevalo \bgroup et al\mbox.\egroup
  }{2017}]{Arevalo:2017iclr}
Arevalo, J.; Solorio, T.; Montes-y G\'omez, M.; and Gonz\'alez, F.~A.
\newblock 2017.
\newblock Gated multimodal units for information fusion.

\bibitem[\protect\citeauthoryear{Atrey \bgroup et al\mbox.\egroup
  }{2010}]{Atrey:2010ms}
Atrey, P.~K.; Hossain, M.~A.; El~Saddik, A.; and Kankanhalli, M.~S.
\newblock 2010.
\newblock Multimodal fusion for multimedia analysis: a survey.
\newblock {\em Multimedia systems} 16(6):345--379.

\bibitem[\protect\citeauthoryear{Baker, Kiela, and
  Korhonen}{2016}]{Baker:2016coling}
Baker, S.; Kiela, D.; and Korhonen, A.
\newblock 2016.
\newblock Robust text classification for sparsely labelled data using
  multi-level embeddings.
\newblock In {\em Proceedings of COLING},  2333--2343.

\bibitem[\protect\citeauthoryear{Baroni}{2016}]{Baroni:2016llc}
Baroni, M.
\newblock 2016.
\newblock Grounding distributional semantics in the visual world.
\newblock {\em Language and Linguistics Compass} 10(1):3--13.

\bibitem[\protect\citeauthoryear{Bernardi \bgroup et al\mbox.\egroup
  }{2016}]{Bernardi:2016jair}
Bernardi, R.; Cakici, R.; Elliott, D.; Erdem, A.; Erdem, E.; Ikizler-Cinbis,
  N.; Keller, F.; Muscat, A.; and Plank, B.
\newblock 2016.
\newblock Automatic description generation from images: A survey of models,
  datasets, and evaluation measures.
\newblock {\em J. Artif. Intell. Res.(JAIR)} 55:409--442.

\bibitem[\protect\citeauthoryear{Borko and Bernick}{1963}]{Borko:1963acm}
Borko, H., and Bernick, M.
\newblock 1963.
\newblock Automatic document classification.
\newblock {\em Journal of the ACM} 10(2):151--162.

\bibitem[\protect\citeauthoryear{Branavan, Silver, and
  Barzilay}{2012}]{Branavan:2012jair}
Branavan, S.; Silver, D.; and Barzilay, R.
\newblock 2012.
\newblock Learning to win by reading manuals in a monte-carlo framework.
\newblock {\em Journal of Artificial Intelligence Research} 43:661--704.

\bibitem[\protect\citeauthoryear{Bruni, Tran, and
  Baroni}{2014}]{Bruni:2014jair}
Bruni, E.; Tran, N.-K.; and Baroni, M.
\newblock 2014.
\newblock Multimodal distributional semantics.
\newblock {\em J. Artif. Intell. Res.(JAIR)} 49(2014):1--47.

\bibitem[\protect\citeauthoryear{Chen \bgroup et al\mbox.\egroup
  }{2009}]{Chen:2009esa}
Chen, J.; Huang, H.; Tian, S.; and Qu, Y.
\newblock 2009.
\newblock Feature selection for text classification with na{\"\i}ve bayes.
\newblock {\em Expert Systems with Applications} 36(3):5432--5435.

\bibitem[\protect\citeauthoryear{Deerwester \bgroup et al\mbox.\egroup
  }{1990}]{Deerwester:1990jasis}
Deerwester, S.; Dumais, S.~T.; Furnas, G.~W.; Landauer, T.~K.; and Harshman, R.
\newblock 1990.
\newblock Indexing by latent semantic analysis.
\newblock {\em Journal of the American society for information science}
  41(6):391.

\bibitem[\protect\citeauthoryear{Frome \bgroup et al\mbox.\egroup
  }{2013}]{Frome:2013nips}
Frome, A.; Corrado, G.~S.; Shlens, J.; Bengio, S.; Dean, J.; Mikolov, T.;
  et~al.
\newblock 2013.
\newblock Devise: A deep visual-semantic embedding model.
\newblock In {\em Advances in neural information processing systems},
  2121--2129.

\bibitem[\protect\citeauthoryear{Fukui \bgroup et al\mbox.\egroup
  }{2016}]{Fukui:2016arxiv}
Fukui, A.; Park, D.~H.; Yang, D.; Rohrbach, A.; Darrell, T.; and Rohrbach, M.
\newblock 2016.
\newblock Multimodal compact bilinear pooling for visual question answering and
  visual grounding.
\newblock {\em CoRR} abs/1606.01847.

\bibitem[\protect\citeauthoryear{He \bgroup et al\mbox.\egroup
  }{2016}]{He:2016cvpr}
He, K.; Zhang, X.; Ren, S.; and Sun, J.
\newblock 2016.
\newblock Deep residual learning for image recognition.
\newblock In {\em Proceedings of the IEEE Conference on Computer Vision and
  Pattern Recognition},  770--778.

\bibitem[\protect\citeauthoryear{Hill, Cho, and Korhonen}{2016}]{Hill:2016iclr}
Hill, F.; Cho, K.; and Korhonen, A.
\newblock 2016.
\newblock Learning distributed representations of sentences from unlabelled
  data.
\newblock {\em arXiv preprint arXiv:1602.03483}.

\bibitem[\protect\citeauthoryear{Jegou, Douze, and
  Schmid}{2011}]{Jegou:2011pami}
Jegou, H.; Douze, M.; and Schmid, C.
\newblock 2011.
\newblock Product quantization for nearest neighbor search.
\newblock {\em IEEE Transactions on Pattern Analysis and Machine Intelligence}
  33(1):117--128.

\bibitem[\protect\citeauthoryear{Joachims}{1998}]{Joachims:1998ecml}
Joachims, T.
\newblock 1998.
\newblock Text categorization with support vector machines: Learning with many
  relevant features.
\newblock In {\em European Conference on Machine Learning},  137--142.

\bibitem[\protect\citeauthoryear{Joulin \bgroup et al\mbox.\egroup
  }{2016}]{Joulin:2016arxiv}
Joulin, A.; Grave, E.; Bojanowski, P.; and Mikolov, T.
\newblock 2016.
\newblock Bag of tricks for efficient text classification.
\newblock {\em CoRR} abs/1607.01759.

\bibitem[\protect\citeauthoryear{Kiela and Bottou}{2014}]{Kiela:2014emnlp}
Kiela, D., and Bottou, L.
\newblock 2014.
\newblock Learning image embeddings using convolutional neural networks for
  improved multi-modal semantics.
\newblock In {\em EMNLP},  36--45.

\bibitem[\protect\citeauthoryear{Kiela and Clark}{2015}]{Kiela:2015emnlp}
Kiela, D., and Clark, S.
\newblock 2015.
\newblock Multi-and cross-modal semantics beyond vision: Grounding in auditory
  perception.
\newblock In {\em EMNLP},  2461--2470.

\bibitem[\protect\citeauthoryear{Kiela, Bulat, and Clark}{2015}]{Kiela:2015acl}
Kiela, D.; Bulat, L.; and Clark, S.
\newblock 2015.
\newblock Grounding semantics in olfactory perception.
\newblock In {\em ACL (2)},  231--236.

\bibitem[\protect\citeauthoryear{Kim}{2014}]{Kim:2014iclr}
Kim, Y.
\newblock 2014.
\newblock Convolutional neural networks for sentence classification.
\newblock In {\em Proceedings of EMNLP}.

\bibitem[\protect\citeauthoryear{Kiros \bgroup et al\mbox.\egroup
  }{2015}]{Kiros:2015nips}
Kiros, R.; Zhu, Y.; Salakhutdinov, R.; Zemel, R.~S.; Torralba, A.; Urtasun, R.;
  and Fidler, S.
\newblock 2015.
\newblock Skip-thought vectors.
\newblock In {\em Proceedings of NIPS}.

\bibitem[\protect\citeauthoryear{Kiros, Salakhutdinov, and
  Zemel}{2014}]{Kiros:2014icml}
Kiros, R.; Salakhutdinov, R.; and Zemel, R.~S.
\newblock 2014.
\newblock Multimodal neural language models.
\newblock In {\em Proceedings of {ICML}}, volume~14,  595--603.

\bibitem[\protect\citeauthoryear{Lazaridou, Bruni, and
  Baroni}{2014}]{Lazaridou:2014acl}
Lazaridou, A.; Bruni, E.; and Baroni, M.
\newblock 2014.
\newblock Is this a wampimuk? cross-modal mapping between distributional
  semantics and the visual world.
\newblock In {\em ACL (1)},  1403--1414.

\bibitem[\protect\citeauthoryear{Lazaridou, Pham, and
  Baroni}{2015}]{Lazaridou:2015iclr}
Lazaridou, A.; Pham, N.~T.; and Baroni, M.
\newblock 2015.
\newblock Combining language and vision with a multimodal skip-gram model.
\newblock {\em arXiv preprint arXiv:1501.02598}.

\bibitem[\protect\citeauthoryear{Le and Mikolov}{2014}]{Le:2014icml}
Le, Q.~V., and Mikolov, T.
\newblock 2014.
\newblock Distributed representations of sentences and documents.
\newblock In {\em ICML}, volume~14,  1188--1196.

\bibitem[\protect\citeauthoryear{Lopopolo and van
  Miltenburg}{2015}]{Lopopolo:2015iwcs}
Lopopolo, A., and van Miltenburg, E.
\newblock 2015.
\newblock Sound-based distributional models.
\newblock {\em IWCS 2015} ~70.

\bibitem[\protect\citeauthoryear{Mei, Bansal, and Walter}{2016}]{Mei:2016aaai}
Mei, H.; Bansal, M.; and Walter, M.~R.
\newblock 2016.
\newblock Listen, attend, and walk: Neural mapping of navigational instructions
  to action sequences.
\newblock In {\em Proceedings of AAAI}.

\bibitem[\protect\citeauthoryear{Mikolov \bgroup et al\mbox.\egroup
  }{2013}]{Mikolov:2013nips}
Mikolov, T.; Sutskever, I.; Chen, K.; Corrado, G.~S.; and Dean, J.
\newblock 2013.
\newblock Distributed representations of words and phrases and their
  compositionality.
\newblock In {\em Advances in neural information processing systems},
  3111--3119.

\bibitem[\protect\citeauthoryear{Mooney}{2008}]{Mooney:2008aaai}
Mooney, R.~J.
\newblock 2008.
\newblock Learning to connect language and perception.
\newblock In {\em Proceedings of AAAI}.

\bibitem[\protect\citeauthoryear{Narasimhan, Kulkarni, and
  Barzilay}{2015}]{Narasimhan:2015emnlp}
Narasimhan, K.; Kulkarni, T.; and Barzilay, R.
\newblock 2015.
\newblock Language understanding for text-based games using deep reinforcement
  learning.
\newblock In {\em Proceedings of EMNLP}.

\bibitem[\protect\citeauthoryear{Ngiam \bgroup et al\mbox.\egroup
  }{2011}]{Ngiam:2011icml}
Ngiam, J.; Khosla, A.; Kim, M.; Nam, J.; Lee, H.; and Ng, A.~Y.
\newblock 2011.
\newblock Multimodal deep learning.
\newblock In {\em Proceedings of ICML},  689--696.

\bibitem[\protect\citeauthoryear{Oquab \bgroup et al\mbox.\egroup
  }{2014}]{Oquab:2014cvpr}
Oquab, M.; Bottou, L.; Laptev, I.; and Sivic, J.
\newblock 2014.
\newblock Learning and transferring mid-level image representations using
  convolutional neural networks.
\newblock In {\em Proceedings of the IEEE conference on computer vision and
  pattern recognition},  1717--1724.

\bibitem[\protect\citeauthoryear{Pang and Lee}{2008}]{Pang:2008ftir}
Pang, B., and Lee, L.
\newblock 2008.
\newblock Opinion mining and sentiment analysis.
\newblock {\em Foundations and trends in information retrieval} 2(1-2):1--135.

\bibitem[\protect\citeauthoryear{Perronnin and
  Larlus}{2015}]{Perronnin:2015cvpr}
Perronnin, F., and Larlus, D.
\newblock 2015.
\newblock Fisher vectors meet neural networks: A hybrid classification
  architecture.
\newblock In {\em Proceedings of the IEEE conference on computer vision and
  pattern recognition},  3743--3752.

\bibitem[\protect\citeauthoryear{Potamianos \bgroup et al\mbox.\egroup
  }{2003}]{Potamianos:2003ieee}
Potamianos, G.; Neti, C.; Gravier, G.; Garg, A.; and Senior, A.~W.
\newblock 2003.
\newblock Recent advances in the automatic recognition of audiovisual speech.
\newblock {\em Proceedings of the IEEE} 91(9):1306--1326.

\bibitem[\protect\citeauthoryear{Razavian \bgroup et al\mbox.\egroup
  }{2014}]{Razavian:2014cvpr}
Razavian, A.~S.; Azizpour, H.; Sullivan, J.; and Carlsson, S.
\newblock 2014.
\newblock Cnn features off-the-shelf: an astounding baseline for recognition.
\newblock In {\em Proceedings of the IEEE Conference on Computer Vision and
  Pattern Recognition Workshops},  806--813.

\bibitem[\protect\citeauthoryear{Regneri \bgroup et al\mbox.\egroup
  }{2013}]{Regneri:2013tacl}
Regneri, M.; Rohrbach, M.; Wetzel, D.; Thater, S.; Schiele, B.; and Pinkal, M.
\newblock 2013.
\newblock Grounding action descriptions in videos.
\newblock {\em Transactions of the Association for Computational Linguistics}
  1:25--36.

\bibitem[\protect\citeauthoryear{Sebastiani}{2002}]{Sebastiani:2002acm}
Sebastiani, F.
\newblock 2002.
\newblock Machine learning in automated text categorization.
\newblock {\em ACM Computing Surveys} 34(1):1--47.

\bibitem[\protect\citeauthoryear{Socher \bgroup et al\mbox.\egroup
  }{2011}]{Socher:2011emnlp}
Socher, R.; Pennington, J.; Huang, E.~H.; Ng, A.~Y.; and Manning, C.~D.
\newblock 2011.
\newblock Semi-supervised recursive autoencoders for predicting sentiment
  distributions.
\newblock In {\em Proceedings of the conference on empirical methods in natural
  language processing},  151--161.
\newblock Association for Computational Linguistics.

\bibitem[\protect\citeauthoryear{Socher \bgroup et al\mbox.\egroup
  }{2013}]{Socher:2013nips}
Socher, R.; Ganjoo, M.; Manning, C.~D.; and Ng, A.
\newblock 2013.
\newblock Zero-shot learning through cross-modal transfer.
\newblock In {\em Advances in neural information processing systems},
  935--943.

\bibitem[\protect\citeauthoryear{Sriram \bgroup et al\mbox.\egroup
  }{2010}]{Sriram:2010sigir}
Sriram, B.; Fuhry, D.; Demir, E.; Ferhatosmanoglu, H.; and Demirbas, M.
\newblock 2010.
\newblock Short text classification in twitter to improve information
  filtering.
\newblock In {\em Proceedings of the 33rd International ACM SIGIR Conference on
  Research and Development in Information Retrieval}, SIGIR '10,  841--842.
\newblock New York, NY, USA: ACM.

\bibitem[\protect\citeauthoryear{Srivastava and
  Salakhutdinov}{2012}]{Srivastava:2012nips}
Srivastava, N., and Salakhutdinov, R.~R.
\newblock 2012.
\newblock Multimodal learning with deep boltzmann machines.
\newblock In {\em Advances in neural information processing systems},
  2222--2230.

\bibitem[\protect\citeauthoryear{Thomee \bgroup et al\mbox.\egroup
  }{2016}]{Thomee:2016acm}
Thomee, B.; Shamma, D.; Friedland, G.; Elizalde, B.; Ni, K.; Poland, D.; Borth,
  D.; and Li, L.
\newblock 2016.
\newblock {YFCC100M: The New Data in Multimedia Research}.
\newblock {\em {Communications of the ACM}} 59(2):64--73.

\bibitem[\protect\citeauthoryear{Wang and Manning}{2012}]{Wang:2012acl}
Wang, S., and Manning, C.~D.
\newblock 2012.
\newblock Baselines and bigrams: Simple, good sentiment and topic
  classification.
\newblock In {\em Proceedings of the 50th Annual Meeting of the Association for
  Computational Linguistics: Short Papers-Volume 2},  90--94.
\newblock Association for Computational Linguistics.

\bibitem[\protect\citeauthoryear{Wang \bgroup et al\mbox.\egroup
  }{2015}]{Wang:2015icme}
Wang, X.; Kumar, D.; Thome, N.; Cord, M.; and Precioso, F.
\newblock 2015.
\newblock Recipe recognition with large multimodal food dataset.
\newblock In {\em Workshop CEA of the IEEE International Conference on
  Multimedia \& Exposition (ICME)},  1--6.
\newblock IEEE.

\bibitem[\protect\citeauthoryear{Weston, Bengio, and
  Usunier}{2011}]{Weston:2011ijcai}
Weston, J.; Bengio, S.; and Usunier, N.
\newblock 2011.
\newblock Wsabie: Scaling up to large vocabulary image annotation.
\newblock In {\em IJCAI}, volume~11,  2764--2770.

\bibitem[\protect\citeauthoryear{Wu \bgroup et al\mbox.\egroup
  }{2004}]{Wu:2004acm}
Wu, Y.; Chang, E.~Y.; Chang, K. C.-C.; and Smith, J.~R.
\newblock 2004.
\newblock Optimal multimodal fusion for multimedia data analysis.
\newblock In {\em Proceedings of the 12th annual ACM international conference
  on Multimedia},  572--579.
\newblock ACM.

\bibitem[\protect\citeauthoryear{Xiong and Svensson}{2002}]{Xiong:2002if}
Xiong, N., and Svensson, P.
\newblock 2002.
\newblock Multi-sensor management for information fusion: issues and
  approaches.
\newblock {\em Information fusion} 3(2):163--186.

\bibitem[\protect\citeauthoryear{Zhao \bgroup et al\mbox.\egroup
  }{2003}]{Zhao:2003csur}
Zhao, W.; Chellappa, R.; Phillips, P.~J.; and Rosenfeld, A.
\newblock 2003.
\newblock Face recognition: A literature survey.
\newblock {\em ACM computing surveys (CSUR)} 35(4):399--458.

\end{thebibliography}
\bibliographystyle{aaai}

\end{document}